\documentclass[letterpaper, 10 pt, conference]{ieeeconf}  %

\IEEEoverridecommandlockouts                              %

\usepackage{graphicx}
\usepackage{amsmath,amssymb}
\usepackage{booktabs}
\usepackage{multirow}
\usepackage{xcolor}
\usepackage{cite}
\usepackage{tikz}
\usetikzlibrary{positioning,arrows.meta}
\usepackage{pgfplots}
\pgfplotsset{compat=1.18}  %
\usepgfplotslibrary{groupplots}
\graphicspath{{figures/}}

\title{\LARGE \bf Where Do the Gaussians Go? Foreground Attraction for\\ Camera-Based Weakly Supervised 3D Occupancy}

\author{Jeagon Song$^{1}$ and Jhonghyun An$^{1,*}$%
\thanks{This work has been submitted to the IEEE for possible publication. Copyright may be transferred without notice, after which this version may no longer be accessible.}%
\thanks{$^{1}$School of Computing, Gachon University, Seongnam 13120, Republic of Korea.}%
\thanks{$^{*}$Corresponding author ({\tt\small jhonghyun@gachon.ac.kr}).}%
}

\begin{document}

\maketitle
\thispagestyle{empty}
\pagestyle{empty}

\begin{abstract}
Camera-based 3D semantic occupancy models can be trained without 3D labels by rendering their predictions into the images and supervising them with 2D pseudo labels.
In sparse Gaussian representations trained this way, Gaussian positions are learned only through pixel-wise rendering losses, which we hypothesize give small foreground objects little weight. %
In trained GaussianFlowOcc models, the Gaussian density on foreground voxels is on average only 1.62 times that on background voxels. %
We propose a training-only foreground attraction loss that back-projects foreground pseudo-label pixels with pseudo depth and pulls nearby Gaussians toward these 3D points, favoring Gaussians that already predict the target class; the inference network is unchanged. %
On Occ3D-nuScenes, the loss raises this concentration to 2.40 over three seeds. %
Over three seeds, the mean foreground IoU increases from 6.29 to 6.96 and is higher for each seed, whereas the change in mean mIoU (13.53 to 13.82) is within the seed-to-seed variation of the baseline. %
Single-run ablations further indicate that pseudo-label targets outperform targets placed at ground-truth voxels and that acting on the Gaussian positions alone is not sufficient, and a single-seed test shows that the model degrades less when fewer Gaussians are kept at inference. %

\end{abstract}

\section{Introduction}\label{sec:intro}

Camera-based 3D semantic occupancy prediction describes the surroundings of a vehicle as a voxel grid in which every cell is labeled as free or as one of several semantic classes, providing a dense representation for planning in autonomous driving~\cite{tian2023occ3d}.
Dense 3D occupancy labels, however, are costly to produce, as they rely on additional data such as LiDAR~\cite{boeder2025gaussianflowocc}.
This has motivated methods that learn occupancy from 2D signals alone, by rendering the predicted 3D scene into the cameras and supervising the renderings with 2D labels, pseudo labels, photometric consistency, or features of 2D foundation models~\cite{pan2024renderocc,huang2024selfocc,zhang2025occnerf,jiang2025gausstr,gan2025gaussianocc,boeder2025gaussianflowocc}.

Gaussian-based occupancy models replace the dense voxel grid with a sparse set of 3D Gaussians~\cite{huang2024gaussianformer,boeder2025gaussianflowocc}.
GaussianFlowOcc~\cite{boeder2025gaussianflowocc} is a representative weakly supervised example: it predicts a fixed number of Gaussians with temporal flow from multi-view images and trains them with semantic pseudo labels from Grounded SAM~\cite{ren2024groundedsam} and pseudo depth from Metric3D~\cite{yin2023metric3d}. %
Because the number of Gaussians is fixed, where the network places them determines how much of the representation is spent on each part of the scene.

We argue that rendering supervision provides only a weak signal for placing Gaussians on foreground objects such as vehicles, pedestrians, and barriers. %
Rendering losses are computed per pixel, so an object that covers few pixels contributes little to the loss, while Gaussians on large surfaces such as roads and buildings affect many pixels. %
We find that averaged over three trained GaussianFlowOcc models, foreground voxels receive 132.6 Gaussians per validation frame, compared with 1193.7 for background voxels, and that the Gaussian density per volume on the foreground is only 1.62 times that on the background. %

\begin{figure}[t]
  \centering
  \includegraphics[width=\columnwidth]{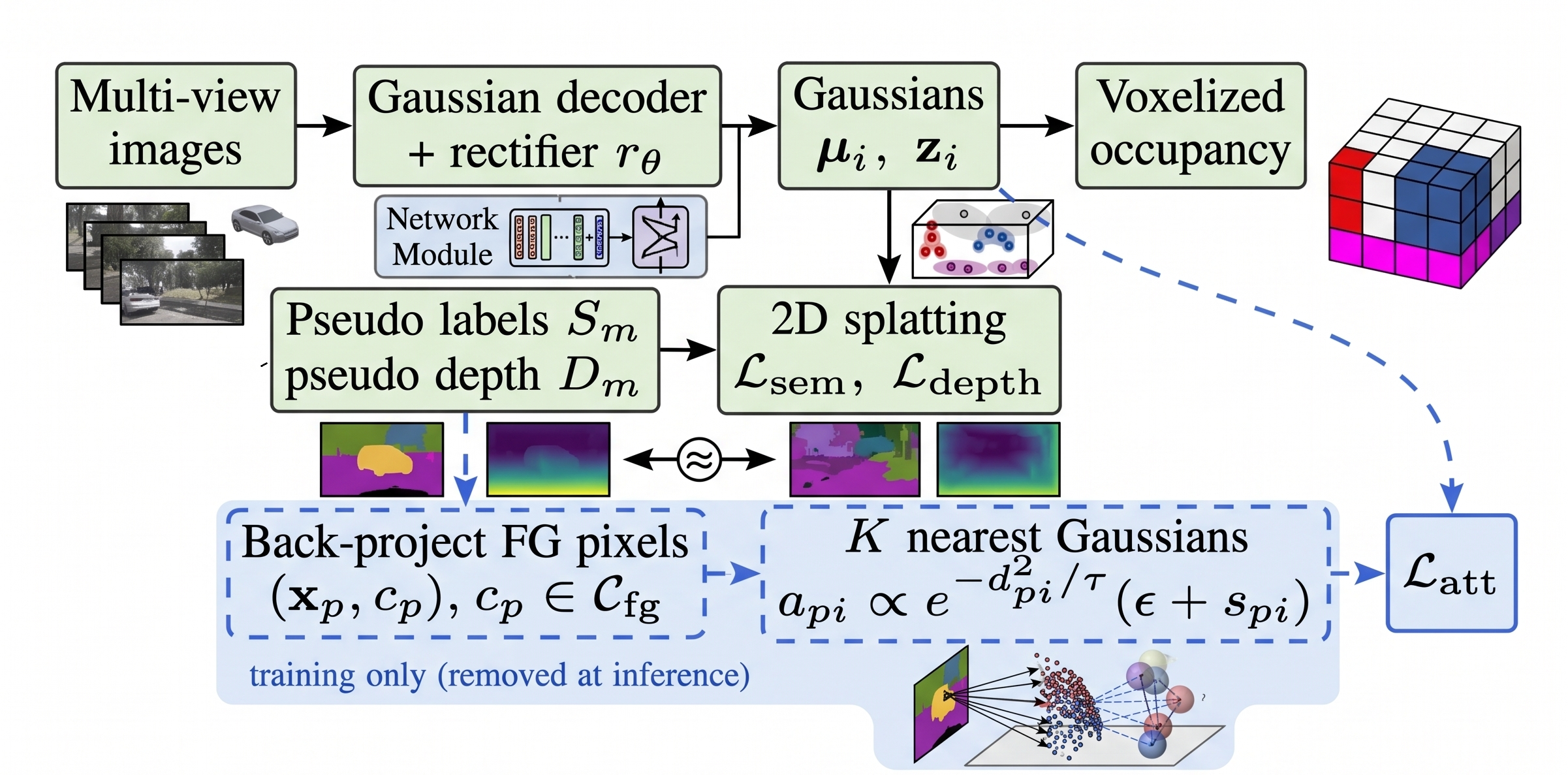}
  \caption{Overview. The baseline learns Gaussian positions only through 2D rendering losses on pseudo semantic labels and pseudo depth. Our foreground attraction loss (dashed, training only) back-projects foreground pseudo-label pixels to 3D and pulls nearby Gaussians toward them, favoring Gaussians that already predict the target class. The inference network is unchanged. Image thumbnails are schematic illustrations.} %
  \label{fig:overview}
\end{figure}

Existing ways to control the distribution of Gaussians do not directly address this setting.
3D Gaussian splatting interleaves optimization and density control for each captured scene~\cite{kerbl20233dgs}, and F4Splat predicts where to densify in a feed-forward manner with an explicit control of the Gaussian budget~\cite{kim2026f4splat}; F4Splat targets novel-view synthesis rather than occupancy prediction.
GaussianFormer-2 observes that many Gaussians in occupancy models describe empty space and initializes them in non-empty regions~\cite{huang2025gaussianformer2}.
Simply adding Gaussians also appears to be of limited use in our setting: doubling their number changed mIoU by less than 0.2 points in two single-run tests with attraction-trained models, although we did not repeat this test for the baseline (Sec.~\ref{sec:ablation}). %
We instead keep the architecture and the number of Gaussians unchanged and add a training-only 3D objective: foreground pixels of the existing pseudo labels are back-projected with the pseudo depth, and nearby Gaussians that already predict the corresponding class are pulled toward them (Fig.~\ref{fig:overview}). %

Our contributions are as follows:
\begin{itemize}
  \item We quantify how Gaussians are distributed between foreground and background under rendering supervision by assigning each Gaussian to the ground-truth voxel containing its center, and show that the resulting foreground concentration is weak (Sec.~\ref{sec:placement}). %
  \item We propose a foreground attraction loss that derives its targets only from the pseudo labels of the baseline, combines distance- and class-aware soft assignment with a robust distance, and leaves the inference network and its cost unchanged (Sec.~\ref{sec:method}). %
  \item On Occ3D-nuScenes, we show over three seeds that the loss increases the foreground concentration from 1.62 to 2.40 and raises the mean foreground IoU while the change in mIoU stays within seed variation; we further analyze the target source, the gradient path, and robustness to a reduced inference budget (Sec.~\ref{sec:experiments}). %
\end{itemize}

\section{Related Work}\label{sec:related}

\subsection{Camera-Based 3D Semantic Occupancy}

Occ3D~\cite{tian2023occ3d} builds dense, visibility-aware occupancy labels for nuScenes~\cite{caesar2020nuscenes} and Waymo, establishing the Occ3D-nuScenes and Occ3D-Waymo benchmarks.
Camera-based methods infer occupancy from a single image, as in semantic scene completion with MonoScene~\cite{cao2022monoscene}, or from multiple cameras with tri-perspective-view features in TPVFormer~\cite{huang2023tpvformer} and dual-path transformers in OccFormer~\cite{zhang2023occformer}.
SurroundOcc~\cite{wei2023surroundocc} supervises multi-camera occupancy prediction at multiple scales with dense occupancy labels generated by fusing multi-frame LiDAR scans.
Subsequent work has focused on efficient scene representations.
FlashOcc~\cite{yu2023flashocc} keeps features in bird's-eye view and lifts them to 3D with a channel-to-height transformation, and SparseOcc~\cite{tang2024sparseocc} replaces dense latent volumes with a sparse latent representation to reduce computation.
GaussianFormer~\cite{huang2024gaussianformer} describes the scene with sparse, object-centric semantic Gaussians that are refined from multi-view images and splatted into voxels, and GaussianFormer-2~\cite{huang2025gaussianformer2} improves efficiency with a probabilistic Gaussian superposition and an initialization in non-empty regions.
EmbodiedOcc~\cite{wu2025embodiedocc} keeps a global memory of 3D semantic Gaussians and updates the Gaussians in observed regions online.
Our work builds on such Gaussian representations but addresses where the Gaussians are placed when no 3D labels are available for supervision.

\subsection{Occupancy Learning from 2D Supervision}

A growing line of work avoids dense 3D labels by rendering the 3D prediction into the cameras.
RenderOcc~\cite{pan2024renderocc} volume-renders a NeRF-style volume and trains it with 2D semantic and depth labels.
SelfOcc~\cite{huang2024selfocc} learns occupancy from video by rendering previous and future frames, and OccNeRF~\cite{zhang2025occnerf} renders depth supervised by multi-frame photometric consistency and obtains semantics from an open-vocabulary 2D segmentation model.
GaussianOcc~\cite{gan2025gaussianocc} uses Gaussian splatting for fully self-supervised occupancy estimation without ground-truth poses, and GaussTR~\cite{jiang2025gausstr} aligns splatted Gaussian features with foundation models.
GaussianFlowOcc~\cite{boeder2025gaussianflowocc} combines sparse Gaussians with temporal flow and supervises them with pseudo semantic labels and pseudo depth.
Rendering has also been used as an auxiliary training signal with 3D labels: GaussRender~\cite{chambon2025gaussrender} projects predicted and ground-truth occupancy into camera views with Gaussian splatting and enforces their consistency, leaving inference unchanged.
These methods differ in how the 2D supervision is obtained, but in the weakly and self-supervised ones the 3D prediction is supervised only through rendered 2D views.
We complement this supervision with a 3D objective derived from the same pseudo labels.

\subsection{Controlling the Distribution of Gaussians}

In 3D Gaussian splatting~\cite{kerbl20233dgs}, the set of Gaussians is adapted by density control during the optimization of each scene, and Rota Bul\`o et al.~\cite{rotabulo2024revising} revise this densification with a pixel-error criterion and a limit on the total number of primitives.
Feed-forward methods predict Gaussians directly from images: pixelSplat~\cite{charatan2024pixelsplat} samples Gaussian centers from predicted probability distributions, and MVSplat~\cite{chen2024mvsplat} localizes them with a plane-sweep cost volume.
F4Splat~\cite{kim2026f4splat} predicts densification scores in a feed-forward manner, allocating Gaussians according to spatial complexity and view overlap while allowing explicit control of the final budget.
For occupancy, MR-Occ~\cite{seong2024mrocc} refines features only for critical voxels in a camera-LiDAR setting, and GaussianFormer-2~\cite{huang2025gaussianformer2} initializes Gaussians in non-empty regions.
A simpler alternative is to reweight the rendering losses by class frequency, which the public GaussianFlowOcc implementation offers as an option; such reweighting changes the contribution of pixels in image space but provides no 3D target for the Gaussian positions. %
In contrast to methods that adapt the number, initialization, or resolution of the representation, we keep the Gaussian budget and the network fixed and change only the training objective so that the learned positions concentrate on foreground objects.

\section{Method}\label{sec:method}

Fig.~\ref{fig:overview} summarizes our approach.
We keep the GaussianFlowOcc architecture and its training losses unchanged~\cite{boeder2025gaussianflowocc} and add a single loss term that is active only during training. %
The term pulls Gaussian centers toward 3D foreground points obtained from the same 2D pseudo labels that supervise the rendering losses. %

\subsection{Preliminaries: Gaussian Occupancy from 2D Supervision}
\label{sec:method_prelim}

GaussianFlowOcc represents the scene around the ego vehicle with $N$ 3D Gaussians and predicts semantic occupancy from multi-view camera images~\cite{boeder2025gaussianflowocc}. %
Each Gaussian $i$ carries a center $\boldsymbol{\mu}_i$ and semantic logits $\mathbf{z}_i \in \mathbb{R}^{C}$ with $C = 16$, among other attributes. %
The decoder consists of $L = 3$ blocks, each of which refines a query feature $\mathbf{f}_i$ through temporal, self-, and cross-attention; each decoder block ends with a rectifier that moves the Gaussian by predicting a residual from its current center: %
\begin{equation}
  \boldsymbol{\mu}_i^{(l)} = \boldsymbol{\mu}_i^{(l-1)} + r_\theta^{(l)}\bigl(\mathbf{f}_i^{(l)}\bigr), \qquad l = 1, \dots, L,
  \label{eq:rectifier}
\end{equation}
where $\boldsymbol{\mu}_i^{(0)}$ is the initial center and $\boldsymbol{\mu}_i = \boldsymbol{\mu}_i^{(L)}$. %
The model is not supervised with 3D occupancy labels. %
Instead, the Gaussians are splatted into each camera, and the rendered semantics and depth are supervised by a semantic loss $\mathcal{L}_{\mathrm{sem}}$ against Grounded SAM pseudo labels~\cite{ren2024groundedsam} and a depth loss $\mathcal{L}_{\mathrm{depth}}$ against Metric3D pseudo depth~\cite{yin2023metric3d}. %
We denote the complete original training objective by $\mathcal{L}_{\mathrm{base}}$.
Consequently, the rectifiers $r_\theta^{(l)}$, which decide where the Gaussians go, are trained only through these image-space losses. %

\subsection{Foreground Concentration under Rendering Supervision}
\label{sec:method_bias}

We hypothesize that rendering supervision gives the rectifiers only a weak incentive to place Gaussians on foreground objects. %
Rendering losses are computed per pixel, so the influence of a scene element grows with its projected area: a pedestrian or a traffic cone covers few pixels, whereas a Gaussian on the road or on a building facade affects many. %
To examine the resulting placement, we assign each Gaussian to the ground-truth voxel that contains its center; ground truth is used here only for analysis. %
Averaged over three trained baseline models and all validation frames, 132.6 Gaussians per frame lie in foreground voxels and 1193.7 in background voxels. %
Normalized by the occupied volume of each group, the Gaussian density on the foreground is only 1.62 times that on the background (Sec.~\ref{sec:placement}). %
Rather than adding Gaussians, we aim to raise this concentration by giving the rectifiers a 3D objective whose strength does not depend on projected area.

\subsection{Foreground Targets from 2D Pseudo Labels}
\label{sec:method_targets}

We construct attraction targets from the pseudo labels that already supervise the baseline, so no new supervision source is introduced. %
Let $\mathcal{C}_{\mathrm{fg}}$ denote the set of foreground classes (barrier, bicycle, bus, car, construction vehicle, motorcycle, pedestrian, traffic cone, trailer, and truck). %
For camera $m$ with intrinsic matrix $\mathbf{K}_m$ and extrinsic transform $\mathbf{T}_m$, a pixel $(u,v)$ is back-projected with its pseudo depth $D_m(u,v)$ only if its pseudo label $S_m(u,v)$ belongs to $\mathcal{C}_{\mathrm{fg}}$, all pixels in its $7 \times 7$ neighborhood carry the same label, its pseudo depth is at most 40\,m, and the resulting point lies inside the occupancy grid: %
\begin{equation}
  \begin{bmatrix} \mathbf{x}_p \\ 1 \end{bmatrix}
  = \mathbf{T}_m^{-1}
  \begin{bmatrix} D_m(u,v)\, \mathbf{K}_m^{-1} [u,\, v,\, 1]^{\top} \\ 1 \end{bmatrix},
  \qquad c_p = S_m(u,v).
  \label{eq:backproject}
\end{equation}
The neighborhood condition removes pixels near label boundaries, where monocular depth tends to be blurred across object edges; without it, a foreground label could be paired with the depth of an adjacent surface and back-projected to a wrong location. %
For every training sample, we select $P$ targets from these candidates. %
If there are at most $P$ candidates, all of them are used; otherwise, $P$ candidates are drawn without replacement, each weighted by the inverse of the number of candidates of its class, so that every class present contributes a similar number of targets. %
Because the targets combine two pseudo-label sources, they contain class and depth errors; the loss below is designed to tolerate them. %

\subsection{Foreground Attraction Loss}
\label{sec:method_loss}

For each target $p$, we consider only the set $\mathcal{N}_K(p)$ of the $K$ Gaussians whose centers are closest to $\mathbf{x}_p$, with distances $d_{pi} = \lVert \mathbf{x}_p - \boldsymbol{\mu}_i \rVert_2$. %
A Gaussian should be responsible for the target when it is close and when it already predicts the target class.
We express the second condition with the class probability $s_{pi} = \mathrm{softmax}(\mathbf{z}_i)_{c_p}$ and define the soft assignment %
\begin{equation}
  \tilde{a}_{pi} = \frac{\exp(-d_{pi}^{2}/\tau)}{\sum_{j \in \mathcal{N}_K(p)} \exp(-d_{pj}^{2}/\tau)} \left(\epsilon + s_{pi}\right),
  \label{eq:assignment}
\end{equation}
\begin{equation}
  a_{pi} = \tilde{a}_{pi} \Big/ \textstyle\sum_{j \in \mathcal{N}_K(p)} \tilde{a}_{pj},
  \label{eq:assignment_norm}
\end{equation}
where $\tau$ is a distance temperature and $\epsilon$ is a floor on the class term. %
The attraction loss is the class-weighted average of robust distances over all targets:
\begin{equation}
  \mathcal{L}_{\mathrm{att}} = \gamma(t)\,
  \frac{\sum_{p} w_{c_p} \sum_{i \in \mathcal{N}_K(p)} a_{pi}\, \rho_\delta(d_{pi})}{\sum_{p} w_{c_p}},
  \label{eq:attract}
\end{equation}
where $\rho_\delta$ is the Huber function with threshold $\delta$, %
\begin{equation}
  \rho_\delta(d) =
  \begin{cases}
    \tfrac{1}{2} d^{2}, & d \le \delta, \\
    \delta \left(d - \tfrac{1}{2}\delta\right), & d > \delta.
  \end{cases}
  \label{eq:huber}
\end{equation}
Without the class term and the class weights, Eq.~\eqref{eq:attract} reduces to a soft, one-sided Chamfer-type distance from the target points to the Gaussian centers~\cite{fan2017pointset}; the class term restricts which Gaussians are attracted.
The class weight is a normalized log inverse frequency,
\begin{equation}
  w_c = \frac{\bar{C} \log(F / F_c)}{\sum_{c'} \log(F / F_{c'})}, \qquad F = \textstyle\sum_{c} F_c,
  \label{eq:classweight}
\end{equation}
where $F_c$ is the voxel frequency of class $c$ and the sums run over the $\bar{C} = 17$ classes of the frequency table, which, unlike the semantic logits, include the \emph{others} class. %
We use the per-class voxel frequencies that are provided with the public implementation of GaussianFlowOcc~\cite{boeder2025gaussianflowocc}; these dataset-level statistics are the only quantity derived from 3D labels that enters training. %
The schedule $\gamma(t)$ is zero for the first $t_{\mathrm{w}}$ iterations, increases linearly to one over the next $t_{\mathrm{r}}$ iterations, and stays at one afterwards. %

\subsection{Design Rationale}
\label{sec:method_design}

Each component of Eqs.~\eqref{eq:assignment}--\eqref{eq:classweight} is motivated by a failure mode that we expect without it; we did not ablate the components individually.
\emph{Local candidates.} Restricting the assignment to $\mathcal{N}_K(p)$ keeps the pull local; attracting all Gaussians would drag the whole scene toward the foreground. %
\emph{Distance term.} The softmax over $-d_{pi}^{2}/\tau$ acts as a soft nearest-neighbor selection, so the one or two closest Gaussians receive most of the assignment weight. %
\emph{Class term.} The factor $\epsilon + s_{pi}$ favors Gaussians that already describe the target class, so the loss is less likely to pull Gaussians away from neighboring background surfaces and leave holes there. The floor $\epsilon$ keeps the assignment defined when no nearby Gaussian predicts the class yet, which is exactly where attraction is most needed. %
\emph{Normalization.} Because $\sum_i a_{pi} = 1$, every target demands the same total attraction, and well-covered targets are not penalized more than poorly covered ones.
\emph{Robust distance.} The targets inherit monocular depth errors, and the Huber function bounds the gradient from far-away outliers that a squared loss would amplify. %
\emph{Warm-up.} The semantic logits are uninformative at the start of training, so $\gamma(t)$ delays attraction until the class term becomes meaningful. %

Gradients of $\mathcal{L}_{\mathrm{att}}$ reach both the centers $\boldsymbol{\mu}_i$, and thus the rectifiers in Eq.~\eqref{eq:rectifier}, and the semantic logits $\mathbf{z}_i$ through $s_{pi}$. %
We keep both paths. In a single-seed ablation with pseudo-label targets, stopping the gradient through $s_{pi}$ reduced mIoU from 13.80 to 12.14, whereas the same change raised mIoU when the targets were taken from 3D ground truth (Sec.~\ref{sec:ablation}). %

\subsection{Training and Inference}
\label{sec:method_train}

The model is trained end to end with
\begin{equation}
  \mathcal{L} = \mathcal{L}_{\mathrm{base}} + \lambda\, \mathcal{L}_{\mathrm{att}},
  \label{eq:total}
\end{equation}
where $\lambda$ balances the attraction term against the rendering losses. %
The targets and $\mathcal{L}_{\mathrm{att}}$ are computed only during training.
The inference network is therefore identical to the baseline by construction, with the same number of Gaussians, parameters, and operations (Sec.~\ref{sec:budget}). %

\section{Experiments}\label{sec:experiments}

\subsection{Setup}
\label{sec:setup}

\textbf{Dataset and metrics.}
We evaluate on Occ3D-nuScenes~\cite{tian2023occ3d,caesar2020nuscenes}, whose 3D occupancy labels are used only for evaluation and analysis. %
Owing to our compute budget, all models are trained on the same 100 scenes, sampled at random from the first 385 of the 700 training scenes (in dataset order), for which pseudo labels had been generated, and evaluated on all 150 validation scenes. %
We evaluate only camera-visible voxels at a resolution of 0.4\,m. %
We report the mean IoU (mIoU) over 15 semantic classes (excluding \emph{others} and \emph{other flat}), the mean IoU over the 10 foreground classes (FG mIoU), and the occupancy IoU, i.e., the IoU of the occupied class in the binary occupied/free classification. %
Because the original GaussianFlowOcc results are obtained with all 700 training scenes~\cite{boeder2025gaussianflowocc}, our numbers are not comparable to those reported in that work; we therefore compare against GaussianFlowOcc trained and evaluated in the same setting. %

\textbf{Pseudo labels.}
Semantic pseudo labels are generated with Grounded SAM~\cite{ren2024groundedsam}, which combines the open-set detector Grounding DINO~\cite{liu2024groundingdino} with the promptable segmentation model SAM~\cite{kirillov2023sam}, in one of two prompting modes: \emph{separate}, which queries each class name individually, and \emph{combine}, which concatenates all class names into one prompt. %
In the combine mode, the class name ``traffic cone'' is split into several tokens and cannot be recovered from the prompt. %
The baseline trained with combine-mode labels obtains zero IoU on construction vehicle, traffic cone, and trailer. %
Our main comparison uses separate-mode labels with three training seeds for both the baseline and our method. %
The ablation in Sec.~\ref{sec:ablation} was run earlier with combine-mode labels, a single run per variant, and a shared seed; we report it separately and do not compare its numbers with the main results. %

\textbf{Implementation details.}
Both methods share the GaussianFlowOcc configuration with $N = 10{,}000$ Gaussians, 18 training epochs, a batch size of 2, and gradient clipping at a norm of 5, trained on a single RTX 5080 GPU with 16\,GB of memory. %
For the attraction loss, we use $K = 8$, $\tau = 1.0\,\mathrm{m}^2$, $\epsilon = 0.1$, $\delta = 1.0$\,m, $\lambda = 0.3$, $P = 512$, and $t_{\mathrm{w}} = t_{\mathrm{r}} = 2000$ iterations. %
These values were set once at design time and were not tuned on the validation set. %
Training the baseline takes approximately 12\,h. %

\subsection{Main Results}
\label{sec:main}

\begin{table}[t]
\centering
\caption{Main comparison on Occ3D-nuScenes (separate-mode pseudo labels, trained on 100 scenes, evaluated on all 150 validation scenes, camera-visible voxels). Mean $\pm$ sample standard deviation over three seeds. mIoU: mean IoU over 15 semantic classes; FG mIoU: mean IoU over the 10 foreground classes; Occ.\ IoU: occupancy IoU.}  %
\label{tab:main3seed}
\small
\setlength{\tabcolsep}{3pt}
\begin{tabular}{lccc}
\toprule
Method & mIoU & FG mIoU & Occ.\ IoU \\
\midrule
GaussianFlowOcc & 13.53 $\pm$ 0.62 & 6.29 $\pm$ 0.50 & 37.72 $\pm$ 1.25 \\  %
Ours & 13.82 $\pm$ 0.10 & 6.96 $\pm$ 0.27 & 37.52 $\pm$ 0.74 \\  %
\bottomrule
\end{tabular}
\end{table}

Table~\ref{tab:main3seed} compares our method with the baseline over three seeds.
Runs with the same seed index share the random seed, so we also compare them pairwise. %
Foreground attraction raises the mean FG mIoU from 6.29 to 6.96, and the FG mIoU is higher for each of the three seeds ($+0.51$, $+0.21$, $+1.30$). %
The mean mIoU changes from 13.53 to 13.82, but per seed index the difference is $-0.12$, $-0.03$, and $+1.03$, and the mean difference of 0.29 is smaller than the standard deviation of the baseline (0.62). %
We therefore do not claim an improvement in mIoU; the FG difference of 0.67 is comparable to the standard deviation of the baseline (0.50), so the results are consistent with, but do not establish, a gain on the foreground classes. %
The seed-to-seed standard deviation of mIoU is lower for our method (0.10 vs.\ 0.62), but three seeds are too few to establish a difference in variance. %
The occupancy IoU is essentially unchanged (37.72 for the baseline and 37.52 for ours; per seed $-0.65$, $-0.68$, and $+0.72$), so the loss does not improve the overall geometry. %

\begin{table*}[t]
\centering
\caption{Per-class IoU on Occ3D-nuScenes validation (camera-visible voxels), averaged over three seeds with separate-mode pseudo labels; same setting as Table~\ref{tab:main3seed}. Many per-class differences are smaller than the seed-to-seed variation. The first ten columns are foreground classes.}
\label{tab:perclass3seed}
\small
\setlength{\tabcolsep}{2.6pt}
\begin{tabular}{lcccccccccccccccc}
\toprule
Method & barr.$\uparrow$ & bicy.$\uparrow$ & bus$\uparrow$ & car$\uparrow$ & c.veh.$\uparrow$ & moto.$\uparrow$ & ped.$\uparrow$ & cone$\uparrow$ & trail.$\uparrow$ & truck$\uparrow$ & drive.$\uparrow$ & sidew.$\uparrow$ & terr.$\uparrow$ & manm.$\uparrow$ & veg.$\uparrow$ & mIoU$\uparrow$ \\
\midrule
GaussianFlowOcc & 5.44 & 2.55 & 9.05 & 14.11 & 1.42 & 6.37 & 7.04 & 6.68 & 0.42 & 9.79 & 57.64 & 28.81 & 31.83 & 11.94 & 9.88 & 13.53 \\  %
\midrule
Ours & 6.24 & 3.67 & 12.99 & 14.25 & 0.84 & 6.88 & 7.77 & 7.59 & 0.31 & 9.05 & 58.26 & 26.92 & 31.25 & 11.92 & 9.45 & 13.82 \\  %
\bottomrule
\end{tabular}
\end{table*}

Table~\ref{tab:perclass3seed} shows where the change comes from.
The IoU increases for seven of the ten foreground classes, with the largest gain on bus ($+3.94$) and gains of $+0.73$ to $+1.12$ on pedestrian, bicycle, barrier, and traffic cone. %
Construction vehicle, trailer, and truck decrease slightly ($-0.58$, $-0.11$, and $-0.74$). %
On the five background classes, the mean IoU decreases by 0.46, mainly on sidewalk ($-1.89$) and vegetation ($-0.43$), while the mean over the foreground classes increases by 0.67. %
On average, the gains on objects are accompanied by a small decrease on background surfaces.

\subsection{Where Do the Gaussians Go?}
\label{sec:placement}

We next check whether the loss changes the placement of Gaussians as intended; since the loss directly pulls Gaussians toward foreground points, this analysis is a sanity check of its mechanism rather than evidence for the accuracy gains.
For each Gaussian, we look up the ground-truth voxel that contains its center and label the Gaussian as foreground or background accordingly, ignoring Gaussians in free voxels, over all validation frames and for the three seeds of each method. %
Table~\ref{tab:density} reports the counts and the number of Gaussians per cubic meter of ground-truth voxels in each group. %

\begin{table}[t]
\centering
\caption{Gaussian placement on all 150 validation scenes (separate-mode pseudo labels; mean $\pm$ sample standard deviation over three seeds). A Gaussian is assigned to the ground-truth voxel containing its center; Gaussians in free voxels are ignored. FG/frame: Gaussians in foreground voxels per frame. FG/BG dens.: Gaussians per m$^3$ of ground-truth foreground/background voxels. Ratio: FG/BG density (concentration).}  %
\label{tab:density}
\footnotesize
\setlength{\tabcolsep}{3pt}
\begin{tabular}{lcccc}
\toprule
Method & FG/frame & FG dens. & BG dens. & Ratio \\  %
\midrule
Baseline & 132.6 $\pm$ 11.9 & 1.91 $\pm$ 0.17 & 1.18 $\pm$ 0.05 & 1.62 $\pm$ 0.21 \\  %
Ours & 228.8 $\pm$ 19.9 & 3.30 $\pm$ 0.29 & 1.38 $\pm$ 0.12 & 2.40 $\pm$ 0.11 \\  %
\bottomrule
\end{tabular}
\end{table}

\begin{figure*}[t]
  \centering
  \includegraphics[width=\textwidth]{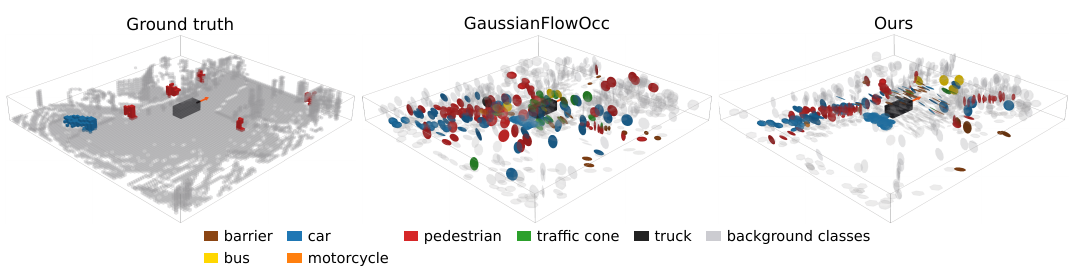}
  \caption{Gaussian placement on a validation frame (scene-0016, frame 8). Left: ground-truth occupancy (camera-visible voxels). Middle: GaussianFlowOcc. Right: ours. Each Gaussian is drawn at its center and scale and colored by its predicted class; background classes are shown in gray. Both models are trained with separate-mode pseudo labels and seed 0.} %
  \label{fig:placement}
\end{figure*}

With the attraction loss, the number of foreground Gaussians per frame increases from 132.6 to 228.8 on average. %
The foreground density rises from 1.91 to 3.30\,m$^{-3}$, whereas the background density changes only from 1.18 to 1.38\,m$^{-3}$. %
The ratio between the two, which we call concentration, grows from 1.62 to 2.40, and every seed of our method has a higher concentration than every baseline seed. %
The number of Gaussians in background voxels also grows (from 1193.7 to 1389.3 per frame), so the additional foreground Gaussians are not taken away from background surfaces. %
On the first 400 validation frames, most Gaussians lie outside the camera-visible voxels that the benchmark evaluates (80.0\,\% for the baseline and 75.9\,\% for ours, averaged over seeds), and a substantial share lies outside the voxel grid (31.6\,\% and 26.1\,\%, the latter varying by 11.0 points across seeds). %
Fig.~\ref{fig:placement} shows an example validation frame rendered with the separate-mode models of Sec.~\ref{sec:main}, in which more of our Gaussians appear near vehicles and pedestrians. %

\subsection{Ablation: Target Source and Gradient Path}
\label{sec:ablation}

\begin{table}[t]
\centering
\caption{Ablation of the attraction target and gradient path (single run per row, same seed, combine-mode pseudo labels; 150 validation scenes). Pos./Cls.: source of the target positions and of the target classes; Detach: stop-gradient through the class term of the assignment. IoU: occupancy IoU. Best in bold.}  %
\label{tab:ablation}
\small
\setlength{\tabcolsep}{3pt}
\begin{tabular}{lccccc}
\toprule
Variant & Pos. & Cls. & Detach & mIoU$\uparrow$ & IoU$\uparrow$ \\
\midrule
No attraction & -- & -- & -- & 13.40 & 38.38 \\  %
\midrule
Pseudo (ours) & pseudo & pseudo & no & \textbf{13.80} & \textbf{39.54} \\  %
Pseudo + detach & pseudo & pseudo & yes & 12.14 & 36.77 \\  %
\midrule
GT & GT & GT & no & 12.91 & 37.80 \\  %
GT + detach & GT & GT & yes & 13.61 & 39.46 \\  %
\midrule
Mix & GT & pseudo & no & 12.21 & 36.38 \\  %
Mix + detach & GT & pseudo & yes & 13.39 & 38.33 \\  %
\bottomrule
\end{tabular}
\end{table}

Table~\ref{tab:ablation} varies where the targets come from and whether the gradient through the class term $s_{pi}$ is stopped; all rows use combine-mode labels and a single run. %
\emph{GT} replaces the pseudo-label targets with the centers and classes of camera-visible Occ3D voxels of the foreground classes, %
 and \emph{Mix} keeps the ground-truth positions but takes the class from the pseudo label of the pixel onto which each voxel projects; if the voxel is occluded in that pixel (depth mismatch above 2\,m), the class term is dropped and the assignment uses distance only. %
Because each variant was trained once, we read these differences as indications rather than conclusive rankings. %

\textbf{Ground-truth target positions do not help by themselves.}
Pseudo-label targets reach 13.80 mIoU, whereas GT and Mix targets reach only 12.91 and 12.21, both below the baseline without attraction (13.40). %
A natural explanation is that ground-truth classes disagree with the pseudo labels that drive the rendering losses: on 300 training frames, 77.7\,\% of the labeled pixels agree with the projected front surface of the Occ3D labels, and the agreement is 0\,\% for construction vehicle, traffic cone, and trailer. %
However, Mix uses the pseudo-label classes and performs worst, so class disagreement alone does not explain the drop. %
Both variants with ground-truth positions also show larger gradients: the median gradient norm over training is 4.72 with pseudo-label targets, 7.32 with GT, and 8.71 with Mix targets, and at the last epoch gradient clipping is active in all iterations of the GT and Mix runs but in none of ours. %
Raising the clipping threshold did not recover the accuracy: resuming the Mix run from epoch 8 with a threshold of 20 yields 12.15 mIoU, compared with 12.21 without this change. %

\textbf{Stopping the class gradient.}
When the gradient through $s_{pi}$ is stopped, the GT and Mix variants recover to 13.61 and 13.39 mIoU, whereas our pseudo-label variant drops from 13.80 to 12.14. %
Stopping the gradient thus helps both variants with ground-truth positions but hurts ours; with single runs, we cannot determine the cause. %
Notably, with the stopped gradient our loss acts only on the positions and its mIoU falls below the baseline (12.14 vs.\ 13.40), suggesting that moving Gaussians toward foreground points is not sufficient by itself and that the effect of the loss on the semantic logits contributes to its benefit. %
Consistent with this reading, placement statistics of the combine-mode models show that this variant still concentrates Gaussians on the foreground (concentration 3.45, compared with 1.49 for the baseline and 4.48 for our model). %
Finally, doubling the number of Gaussians from 10{,}000 to 20{,}000 changed mIoU by only $-0.19$ for our model and $+0.11$ for the GT variant with stopped gradients, suggesting that adding Gaussians is not an effective alternative in this setting, although the baseline was not tested. %

\subsection{Inference Budget and Cost}
\label{sec:budget}

Because the attraction loss is used only for training, our model and the baseline share the same inference network, which has 39.245\,M parameters, 391.48\,G FLOPs, and 13.7 frames per second on an RTX 5080 GPU with a batch size of 1, a $256 \times 704$ input, and 32-bit floating-point precision. %
We further examine how the two models behave when only the $B$ Gaussians with the highest opacity are kept at inference, without retraining. %
This experiment uses the separate-mode models with seed 0 and all 150 validation scenes. %

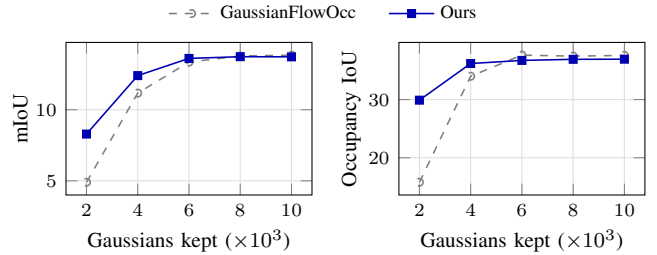
\begin{figure}[t]
  \centering
  \begin{tikzpicture}
\begin{groupplot}[
  group style={group size=2 by 1, horizontal sep=1.15cm},
  width=0.56\columnwidth, height=3.6cm,
  xtick={2,4,6,8,10},
  xlabel={Gaussians kept ($\times 10^3$)},
  tick label style={font=\scriptsize},
  label style={font=\footnotesize},
  xlabel style={yshift=2pt},
  ylabel style={yshift=-4pt},
  grid=major, grid style={gray!25},
  every axis plot/.append style={mark size=1.4pt},
  every axis plot/.append style={semithick},
]
\nextgroupplot[ylabel={mIoU},
  legend to name=budgetlegend,
  legend columns=2,
  legend style={font=\scriptsize, draw=none, /tikz/every even column/.append style={column sep=10pt}},
  legend cell align=left]
\addplot[gray, dashed, mark=o] coordinates {(2,4.90) (4,11.17) (6,13.35) (8,13.78) (10,13.84)}; %
\addplot[blue!70!black, mark=square*] coordinates {(2,8.29) (4,12.40) (6,13.61) (8,13.73) (10,13.72)}; %
\legend{GaussianFlowOcc, Ours}
\nextgroupplot[ylabel={Occupancy IoU}]
\addplot[gray, dashed, mark=o] coordinates {(2,15.79) (4,34.01) (6,37.64) (8,37.47) (10,37.59)}; %
\addplot[blue!70!black, mark=square*] coordinates {(2,29.90) (4,36.20) (6,36.73) (8,36.92) (10,36.94)}; %
\end{groupplot}
\node[anchor=south] at (current bounding box.north) {\pgfplotslegendfromname{budgetlegend}};
\end{tikzpicture}
  \caption{Accuracy when only the $B$ Gaussians with the highest opacity are kept at inference (no retraining). All 150 validation scenes; separate-mode models with seed 0.} %
  \label{fig:budget}
\end{figure}

As shown in Fig.~\ref{fig:budget}, the two models differ little at $B \geq 6{,}000$ (mIoU $-0.12$ to $+0.26$, occupancy IoU $-0.91$ to $-0.55$), but our model degrades much less when the budget is reduced. %
At $B = 4{,}000$, it is higher by $1.23$ mIoU and $2.19$ occupancy IoU, and at $B = 2{,}000$ by $3.39$ and $14.11$. %
This single-seed result suggests that concentrating Gaussians on objects makes the representation more robust to a reduced inference budget.

\subsection{Limitations}
\label{sec:limitations}

Our study has several limitations.
All models are trained on a single GPU, using 100 scenes drawn from the first 385 of the 700 training scenes. So the training data may not represent the full training set, and whether the gains persist with all training scenes remains open. %
With three seeds, the mIoU and occupancy IoU differences are smaller than the variation of the baseline, the ablation relies on single runs with combine-mode pseudo labels, and the budget analysis uses a single seed. %
The evaluation covers only camera-visible voxels, so occluded regions are not assessed. %
The targets inherit pseudo-label errors, and classes that the pseudo labels do not recognize correctly cannot benefit from the loss. %
We did not study the sensitivity to $K$, $\tau$, $\epsilon$, and $\lambda$, did not ablate the individual loss components, and did not compare against reweighting the rendering losses by class frequency.

\section{Conclusion}\label{sec:conclusion}

We studied where a camera-based, weakly supervised Gaussian occupancy model places its Gaussians and found that rendering supervision alone concentrates them only weakly on foreground objects. %
To address this, we introduced a foreground attraction loss that back-projects foreground pseudo labels into 3D targets and pulls nearby Gaussians toward them, favoring those that already predict the target class, without changing the inference network. %
On Occ3D-nuScenes, the loss increased the foreground concentration of Gaussians from 1.62 to 2.40 over three seeds. %
Over three seeds, the mean foreground IoU was higher than that of the baseline, whereas the differences in mIoU and occupancy IoU lie within the seed-to-seed variation of the baseline. %
In a single-seed test, the model also degraded less when fewer Gaussians were kept at inference. %
Single-run ablations showed that targets built from pseudo labels outperformed targets placed at ground-truth voxels and that acting on the positions alone appeared insufficient. %

The improvements are modest relative to seed-to-seed variation, and our experiments use a reduced training set. %
Future work includes training on the full dataset, improving the pseudo labels for classes that are currently not recognized correctly, and adapting the attraction targets to where the rendering losses indicate errors. %

\section*{Acknowledgment}
The authors used Claude (Anthropic), accessed through Claude Code, to assist in drafting the text of all sections, formatting the tables, generating the plotting code for Fig.~\ref{fig:budget}, and checking bibliographic entries.
The illustration in Fig.~\ref{fig:overview} was created by the authors with Gemini (Google), based on outputs of the authors' own pipeline.
All experiments and results were produced by the authors.
The authors reviewed and revised all AI-assisted content and take full responsibility for the paper.

\bibliographystyle{IEEEtran}
\bibliography{refs}

\end{document}